\documentclass[conference]{IEEEtran}
\IEEEoverridecommandlockouts
\usepackage{algorithm}
\usepackage{array}
\usepackage{textcomp}
\usepackage{stfloats}
\usepackage{url}
\usepackage{verbatim}
\usepackage{booktabs}
\usepackage{caption}
\usepackage{amsfonts}
\usepackage{hyperref}
\usepackage{color}
\usepackage{multirow}
\usepackage{makecell}
\usepackage{balance}

\usepackage{cite}
\usepackage{amsmath,amssymb,amsfonts}
\usepackage{algorithmic}
\usepackage{graphicx}
\usepackage{textcomp}
\usepackage{xcolor}
\def\BibTeX{{\rm B\kern-.05em{\sc i\kern-.025em b}\kern-.08em
    T\kern-.1667em\lower.7ex\hbox{E}\kern-.125emX}}
\begin{document}

\title{Knowledge Distillation for Image Restoration : Simultaneous Learning from Degraded and Clean Images\\
\thanks{This work is supported by National Key Research and Development Program of China (No. 2021YFB3101300).}
\thanks{\IEEEauthorrefmark{1} Corresponding author (D. Yan).}
}

\author{
\IEEEauthorblockN{
    Yongheng Zhang,
    and Danfeng Yan\IEEEauthorrefmark{1}}
    \IEEEauthorblockA{State Key Laboratory of Networking and Switching Technology, 
    BUPT, Beijing, China\\Email: {zhangyongheng, yandf}@bupt.edu.cn}
}

\maketitle

\begin{abstract}
Model compression through knowledge distillation has seen extensive application in classification and segmentation tasks. 
However, its potential in image-to-image translation, particularly in image restoration, remains underexplored.
To address this gap, we propose a \textbf{S}imultaneous \textbf{L}earning \textbf{K}nowledge \textbf{D}istillation (\textbf{SLKD}) framework tailored for model compression in image restoration tasks.
SLKD employs a dual-teacher, single-student architecture with two distinct learning strategies: \textbf{D}egradation \textbf{R}emoval \textbf{L}earning (\textbf{DRL}) and \textbf{I}mage \textbf{R}econstruction \textbf{L}earning (\textbf{IRL}), simultaneously.
In DRL, the student encoder learns from Teacher A to focus on removing degradation factors, guided by a novel BRISQUE extractor. 
In IRL, the student decoder learns from Teacher B to reconstruct clean images, with the assistance of a proposed PIQE extractor. 
These strategies enable the student to learn from degraded and clean images simultaneously, ensuring high-quality compression of image restoration models.
Experimental results across five datasets and three tasks demonstrate that SLKD achieves substantial reductions in FLOPs and parameters, exceeding 80\%, while maintaining strong image restoration performance.

\end{abstract}
\begin{IEEEkeywords}
Knowledge distillation, image restoration, simultaneous learning, feature extractor
\end{IEEEkeywords}

\section{Introduction}
\label{sec:intro}

Image restoration is vital in outdoor systems with limited computing resources, such as security surveillance and remote sensing satellites, making model compression essential.
The concept of model compression through knowledge distillation was first introduced for image classification tasks \cite{hinton2015distilling} and has since been widely explored in areas like object detection \cite{romero2014fitnets, zagoruyko2016paying, kim2018paraphrasing,yang2019snapshot, zhao2022decoupled, wang2024crosskd}. 
However, unlike the feature extractor and classification projector commonly used in classification tasks, image restoration models typically employ an encoder-decoder architecture \cite{ronneberger2015u}. 
This architectural difference makes it challenging to directly apply knowledge distillation methods designed for classification tasks to image restoration.

Recent advances in knowledge distillation for compressing image transfer models \cite{li2020semantic, jin2021teachers, zhang2022wavelet}, especially in super-resolution \cite{hui2018fast, gao2018image, zhang2021data, fang2023dual}, have shown promise.
However, while super-resolution models primarily focus on reconstructing low-resolution images into high-resolution ones, image restoration models must first remove degradation factors (such as rain, noise, etc.) before reconstructing the image. 
This additional complexity often results in suboptimal performance when using knowledge distillation methods designed for super-resolution in image restoration tasks.

\begin{figure}[t]
\setlength{\abovecaptionskip}{0.1cm}
\setlength{\belowcaptionskip}{-0.4cm}
    \includegraphics[width=\columnwidth]{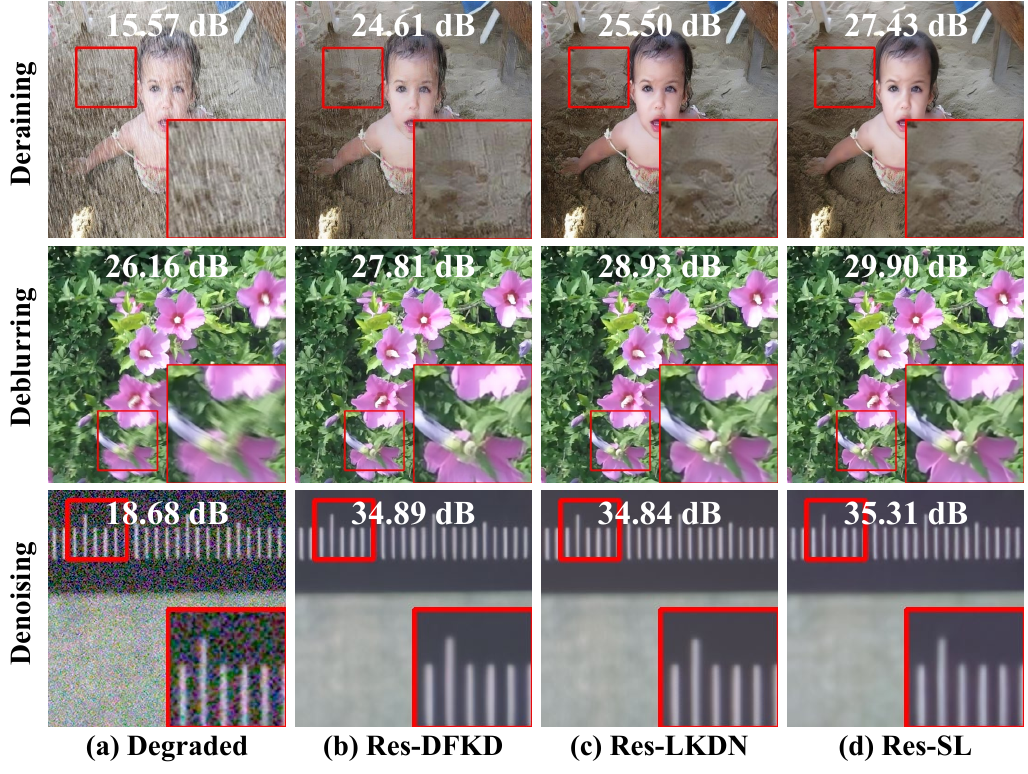}
    \caption{Qualitative comparison with knowledge distillation methods across image deraining, deblurring and denoising.}
  \label{fig:first}
\end{figure}

Building on these insights and the specific requirements of image restoration tasks, we introduce the \textbf{S}imultaneous \textbf{L}earning \textbf{K}nowledge \textbf{D}istillation (\textbf{SLKD}) framework for compressing image restoration models.
For the student encoder, we propose a \textbf{D}egradation \textbf{R}emoval \textbf{L}earning (\textbf{DRL}) strategy, which distills knowledge from Teacher A, a network that processes degraded images.
To ensure the student's focus on degradation removal, DRL incorporates a BRISQUE-based extractor to capture scene statistical features often affected by degradation.
For the student decoder, we introduce an \textbf{I}mage \textbf{R}econstruction \textbf{L}earning (\textbf{IRL}) strategy, which distills knowledge from Teacher B, a network that processes clean images.
In IRL, a PIQE-based extractor is proposed to extract edge and texture features critical for accurate image reconstruction.
These innovations enable SLKD to surpass other knowledge distillation-based model compression methods, as illustrated in Fig. \ref{fig:first}.
Experimental results across five datasets for image deraining, deblurring, and denoising, as detailed in Section \ref{sec:exp}, further demonstrate SLKD's state-of-the-art performance.

\begin{figure*}[t]
\setlength{\abovecaptionskip}{0.3cm}
\setlength{\belowcaptionskip}{-0.3cm}
\centering
  \includegraphics[width=2.0\columnwidth]{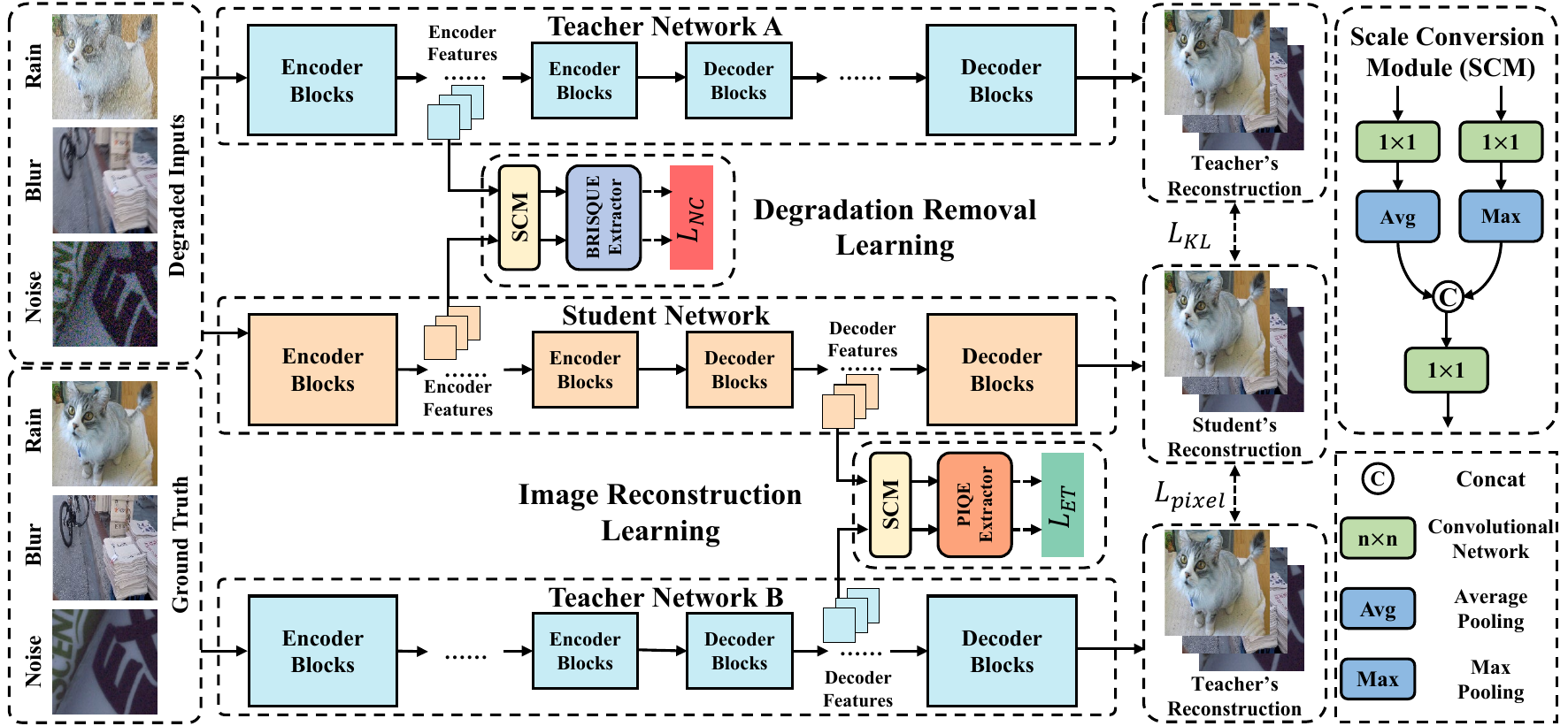}
  \caption{The overall architecture of our proposed Simultaneous Learning Knowledge Distillation (SLKD) strategy for image restoration. SLKD includes a Degradation Removal Learning (DRL) and an Image Reconstruction Learning (IRL).}
  \label{fig:architecture}
\end{figure*}

\section{Proposed Method}
\label{sec:method}
\subsection{Overall Architecture}
As illustrated in Fig. \ref{fig:architecture}, our Simultaneous Learning Knowledge Distillation (SLKD) strategy leverages two pre-trained teacher image restoration networks and a student network, which is trained by distilling knowledge from the teachers.
The student network has a similar structure to the teacher networks but with fewer transformers and reduced feature dimensions in each encoder-decoder block, significantly lowering its complexity.
Given a pair of degraded and clean images $I, G\in\mathcal{R}^{H\times W \times3}$, the teacher network A and the student network both take the degraded image $I$ as input, while the teacher network B uses the clean image $G$.

In the Degradation Removal Learning (DRL) strategy, the student encoder learns from the teacher A encoder, focusing on removing degradation factors and extract clean background features.
This is achieved by mapping encoder features to a unified scale space using a Scale Conversion Module (SCM), followed by a BRISQUE-based extractor to capture natural scene statistics, which are used to compute the natural and color loss $L_{NC}$.

Similarly, in the Image Reconstruction Learning (IRL) strategy, the student decoder learns from the teacher B decoder to reconstruct a clean background image from the extracted features. 
The decoder features are also mapped to a unified scale space, and a PIQE-based extractor is employed to capture edge and texture features, contributing to the edge and texture loss $L_{ET}$. 

In addition to these feature-level learning strategies, we incorporate two image-level losses: KL divergence loss and pixel-level L1 loss.

\subsection{Degradation Removal Learning}
\label{ssec:DRL}
In typical encoder-decoder architectures for image-to-image transfer tasks, the encoder gradually extracts essential features needed for the task. 
In the context of image restoration, these features correspond to a clean feature map, free from degradation factors. 
Thus, we utilize the pre-trained teacher A to restore degraded images, allowing the student encoder to learn how to progressively remove degradation by mimicking the teacher A encoder in the DRL strategy.

In DRL, a SCM first maps the k-th level encoder features from the teacher ($TA_e^k$) and student ($S_e^k$) networks into a unified scale space, resulting in $TA_{eu}^k$ and $S_{eu}^k$.
To ensure the student encoder focuses on removing degradation factors, we introduce a BRISQUE-based extractor to isolate natural and color features that are vulnerable to degradation.
The BRISQUE-based extractor calculates Mean Subtracted Contrast Normalized (MSCN) coefficients of $TA_{eu}^k$ and $S_{eu}^k$ defined as:
{\setlength\abovedisplayskip{5pt}
\setlength\belowdisplayskip{5pt}
\begin{equation}
\begin{aligned}
\hat{I}(i, j) = \frac{I(i, j) - \mu(i, j)}{\sigma(i, j) + C},
\end{aligned}
\end{equation}
}
where $I(i, j)$ represents the pixel value at (i, j), $\mu(i, j)$ and $\sigma(i, j)$ are the local mean and variance, respectively.
The MSCN map is then used to extract natural and color features (${TA}_{ec}^k$ and $S_{ec}^k$) with generalized Gaussian and asymmetric generalized Gaussian distributions, as detailed in \cite{mittal2012no}. 
Finally, the natural and color loss $L_{NC}$ is computed as:
{\setlength\abovedisplayskip{5pt}
\setlength\belowdisplayskip{5pt}
\begin{equation}
\begin{aligned}
L_{NC} = \sum_{k=1}^{n} \sum_{i, j} S_{ec}^k(i, j)*log(\frac{S_{ec}^k(i, j)}{TA_{ec}^k(i, j)}),
\end{aligned}
\end{equation}
}
where $n$ is the number of encoder/decoder blocks.

\subsection{Image Reconstruction Learning}
\label{ssec:IRL}
In the encoder-decoder architecture, the decoder's role is to gradually reconstruct the complete image from extracted features.
For image restoration, this translates to producing a clear background image.
To facilitate this, we use corresponding clean images as input for teacher network B, allowing the student decoder to learn how to restore edge and texture features by mimicking the teacher B decoder in the IRL.

Similarly, a SCM maps the k-th level decoder features from the teacher ($TB_d^k$) and student ($S_d^k$) networks into a unified scale space, resulting in $TB_{du}^k$ and $S_{du}^k$.
To ensure the student decoder focuses on reconstructing clear images, we introduce a PIQE-based extractor to capture crucial edge and texture features for image reconstruction.
The PIQE-based extractor also involves MSCN coefficient calculation described in Equation (1), along with feature vector extraction.
Edge and texture features ($TB_{dc}^k$ and $S_{dc}^k$) are derived using distortion estimation and block variance based on the MSCN coefficients of $TB_{du}^k$ and $S_{du}^k$, as detailed in \cite{venkatanath2015blind}.
The loss $L_{ET}$ is:
{\setlength\abovedisplayskip{5pt}
\setlength\belowdisplayskip{5pt}
\begin{equation}
\begin{aligned}
L_{ET} = \sum_{k=1}^{n} \sum_{i, j} S_{dc}^k(i, j)*log(\frac{S_{dc}^k(i, j)}{TB_{dc}^k(i, j)}),
\end{aligned}
\end{equation}
}

\begin{figure}[t]
\setlength{\abovecaptionskip}{0.1cm}
\setlength{\belowcaptionskip}{-0.2cm}
\centering
  \includegraphics[width=\columnwidth]{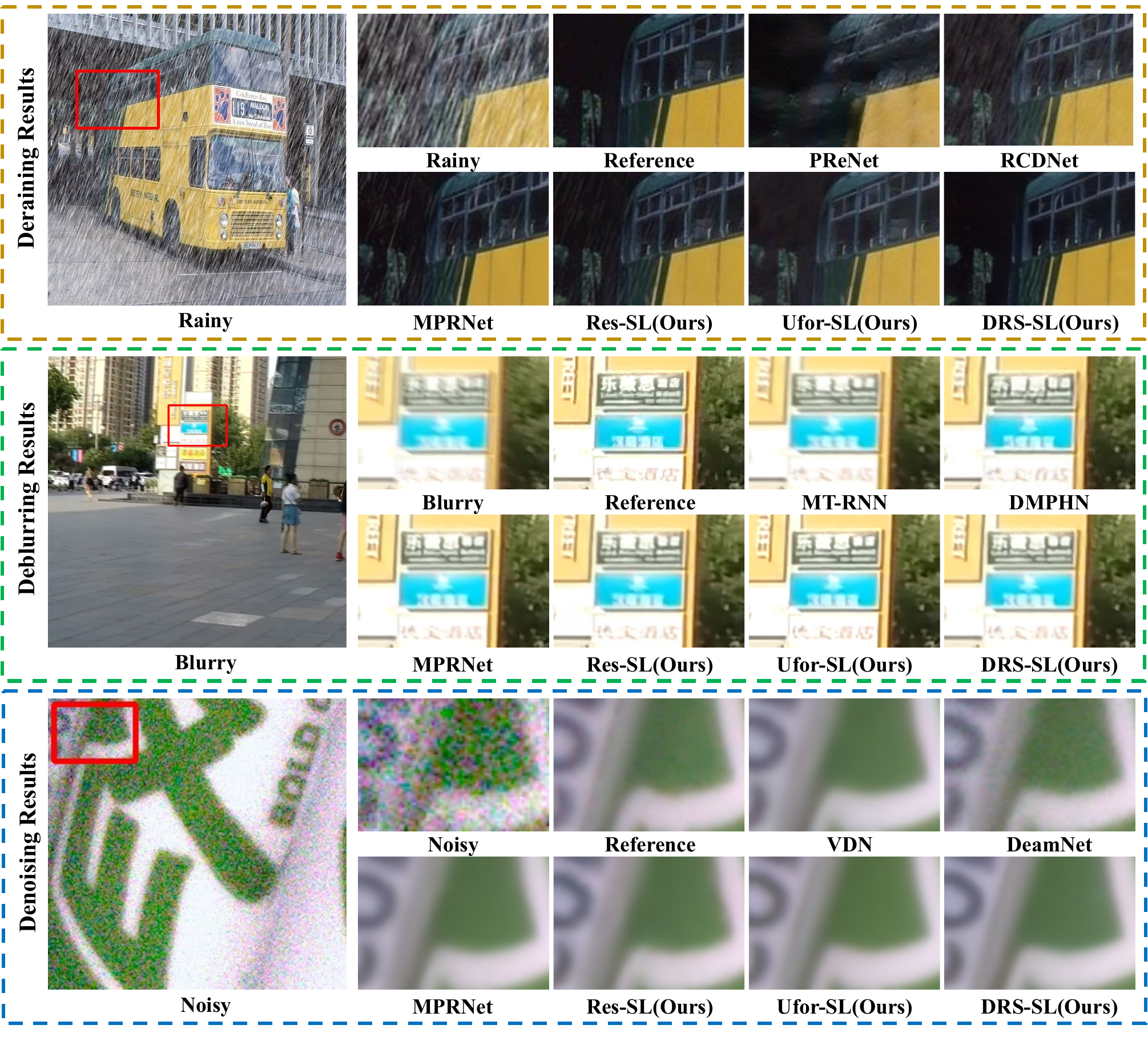}
  \caption{Qualitative comparison with light-weight methods.}
  \label{fig:main}
\end{figure}

\subsection{Overall Loss}
\label{ssec:loss}
The image-level KL loss between student's reconstruction $S_r$ and teacher A's reconstruction $TA_r$ is formulated as:
{\setlength\abovedisplayskip{5pt}
\setlength\belowdisplayskip{5pt}
\begin{equation}
\begin{aligned}
L_{KL} = \sum_{i, j} S_r(i, j)*log(\frac{S_r(i, j)}{TA_r(i, j)}),
\end{aligned}
\end{equation}
}
The pixel-level loss between student's reconstruction $S_r$ and teacher B's reconstruction ${TB}_r$ is formulated as:
{\setlength\abovedisplayskip{5pt}
\setlength\belowdisplayskip{5pt}
\begin{equation}
\begin{aligned}
L_{pixel} = ||S_r - {TB}_r||_1,
\end{aligned}
\end{equation}
}
Finally, the overall loss is formulated as:
{\setlength\abovedisplayskip{5pt}
\setlength\belowdisplayskip{5pt}
\begin{equation}
\begin{aligned}
L_{all} = L_{pixel} + \alpha_1L_{KL} + \alpha_2(L_{NC} + L_{ET}).
\end{aligned}
\end{equation}
}

\begin{table}[tp]
\small
\tabcolsep=5pt
\setlength{\abovecaptionskip}{0.1cm}
\setlength{\belowcaptionskip}{-0cm}
\centering
  \caption{Hyper-parameters and complexity.}
  \label{tab:hyper}
  \begin{tabular}{cccccc}
    \toprule
    Method &    layers  &  dim  & FLOPs & Param. & Infer time \\
    \midrule
    Restormer\cite{zamir2022restormer} &  4,6,6,8 & 48 & 619.5G & 26.1M & 0.1103s\\
    Res-SL  & 1,2,2,4 & 24 & 85.0G & 3.5M & 0.0356s\\
    \midrule
    Uformer\cite{wang2022uformer} & 1,2,8,8 & 32 & 347.6G & 50.9M & 0.1737s\\
    Ufor-SL & 1,2,4,4 & 16 & 57.1G & 7.9M & 0.0540s\\
    \midrule
    DRSformer\cite{chen2023learning} & 4,4,6,6,8 & 48 & 972.0G & 33.7M & 0.3074s\\
    DRS-SL & 2,2,2,2,4 & 24 & 132.0G & 4.6M & 0.0599s\\
    \bottomrule
  \end{tabular}
\end{table}

\begin{table}[tp]
\footnotesize
\renewcommand\arraystretch{1.2}
\tabcolsep=0pt
\setlength{\abovecaptionskip}{0.1cm}
\setlength{\belowcaptionskip}{-0cm}
\centering
  \caption{Quantitative comparison with knowledge distillation methods on five datasets.}
  \label{tab:full-dis}
  \begin{tabular}{c|c|c|c|c|c}
    \toprule
    Tasks &  \multicolumn{2}{c|}{Deraining} & \multicolumn{2}{c|}{Deblurring} & Denoising \\
    \hline
    Datasets & Rain1400\cite{fu2017removing} & Test1200\cite{zhang2018density} & Gopro\cite{nah2017deep} & HIDE\cite{shen2019human} & SIDD\cite{abdelhamed2018high}\\
    \hline
    Restormer\cite{zamir2022restormer} & 34.18/0.944 & 33.19/0.926 & 32.92/0.961 & 31.22/0.942 & 40.02/0.960 \\
    DFKD \cite{zhang2021data} & 32.56/0.928 & 31.95/0.907 & 31.27/0.940 & 29.28/0.923 & 39.23/0.951\\
    DCKD \cite{fang2023dual} & \underline{32.67}/\underline{0.930} & \underline{32.16}/\underline{0.911} & \underline{31.41}/\underline{0.946} & \underline{29.51}/\underline{0.928} & \underline{39.34}/\underline{0.954}\\
    Res-SL &   \textbf{33.24/0.937} & \textbf{32.67/0.917} & \textbf{31.91/0.957} & \textbf{30.17/0.934} & \textbf{39.60/0.957}\\
    \hline
    Uformer\cite{wang2022uformer} & 33.63/0.939 & 33.04/0.923 & 33.06/0.967 & 30.90/0.953 & 39.89/0.960 \\
    DFKD \cite{zhang2021data} & 32.27/0.925 & \underline{32.01}/\underline{0.910} & \underline{31.60}/\underline{0.949} & \underline{29.50}/\underline{0.928} & 39.27/\underline{0.952}\\
    DCKD \cite{fang2023dual} & \underline{32.31}/\underline{0.926} & 31.95/0.909 & 31.52/0.947 & 29.47/0.927 & \underline{39.28}/\underline{0.952}\\
    Ufor-SL &   \textbf{32.96/0.931} & \textbf{32.26/0.915} & \textbf{32.24/0.953} & \textbf{30.35/0.933} & \textbf{39.61/0.956}\\
    \hline
    DRSformer\cite{chen2023learning} & 34.33/0.947 & 33.31/0.927 & 32.76/0.958 & 31.17/0.940 & 40.03/0.960 \\
    DFKD \cite{zhang2021data} & 32.59/0.929 & 32.17/0.911 & 31.21/0.938 & 29.33/0.924 & 39.22/\underline{0.951}\\
    DCKD \cite{fang2023dual} & \underline{32.91}/\underline{0.932} & \underline{32.24}/\underline{0.912} & \underline{31.44}/\underline{0.945} & \underline{29.41}/\underline{0.926} & \underline{39.27}/\underline{0.951}\\
    DRS-SL &  \textbf{33.57/0.938}  & \textbf{32.71/0.917} & \textbf{31.93/0.956} & \textbf{30.26/0.935} & \textbf{39.57/0.956} \\
    \bottomrule
  \end{tabular}
\end{table}

\vspace{-0.3cm}
\section{Experiments and Analysis}
\label{sec:exp}
\noindent\textbf{Datasets and Metrics.} We conduct knowledge distillation experiments using five publicly available datasets: Rain1400\cite{fu2017removing} and Test1200\cite{zhang2018density} for deraining, Gopro\cite{nah2017deep} and HIDE\cite{shen2019human} for deblurring, and SIDD\cite{abdelhamed2018high} for denoising. 
To evaluate restoration performance, we use two full-reference metrics: Peak Signal-to-Noise Ratio (PSNR\cite{huynh2008scope}) in dB and Structural Similarity Index (SSIM\cite{wang2004image}). 
Model complexity is assessed by measuring FLOPs and inference time on each $512 \times 512$ image. 
In the tables, the best and second-best scores for the evaluated methods are highlighted and underlined.

\noindent\textbf{Implementation details.} Our framework is implemented in PyTorch, utilizing the Adam optimizer with parameters ($\beta_{1}$ = 0.9, $\beta_{2}$ = 0.999, weight decay 1e-4). 
The models are trained for 100 epochs, starting with an initial learning rate of 1e-4, which is gradually reduced to 1e-6 using cosine annealing \cite{loshchilov2016sgdr}.
We set the batch size to 8, with a fixed patch size of 128.
The trade-off weights are $\alpha_{1}$ = 0.5 and $\alpha_{2}$ = 0.1.

We select three state-of-the-art restoration methods, Restormer\cite{zamir2022restormer}, Uformer\cite{wang2022uformer}, and DRSformer\cite{chen2023learning}, as teacher models. 
The corresponding student models are named Res-SL, Ufor-SL, and DRS-SL, respectively. 
Table \ref{tab:hyper} shows the hyper-parameters (number of layers and dimensions in each layer of the encoder-decoder) of the teacher models and their corresponding student models, along with the associated reductions in model complexity (FLOPs, inference time, number of parameters) resulting from the reduction in model size.

\begin{table*}[tp]
\small
\tabcolsep=10pt
\renewcommand\arraystretch{1.05}
\setlength{\abovecaptionskip}{0.1cm}
\setlength{\belowcaptionskip}{-0cm}
\centering
  \caption{Quantitative comparison with light-weight methods on five datasets.}
  \label{tab:light-dis}
  \begin{tabular}{cccccccc}
    \toprule
    Methods & Rain1400\cite{fu2017removing} & Test1200\cite{zhang2018density} & Gopro\cite{nah2017deep} & HIDE\cite{shen2019human} & SIDD\cite{abdelhamed2018high} & FLOPs & Infer time \\
    \midrule
    PReNet \cite{ren2019progressive}& 31.75/0.916 & 31.36/0.911 & -/- & -/- & -/- &176.7G&0.0589s\\
    RCDNet \cite{wang2020model}&   32.35/0.926 & 32.13/0.909 & -/- & -/- & -/-&842.5G&0.1919s\\
    \midrule
    DMPHN \cite{zhang2019deep}&  -/-  & -/- & 31.20/0.940 &29.09/0.924&-/-&113.0G&0.0508s\\
    MT-RNN \cite{park2020multi}& -/- & -/- & 31.15/0.945 & 29.15/0.918 & -/- &579.0G&\underline{0.0387s}\\
    \midrule
    VDN\cite{yue2019variational} & -/- & -/- & -/- & -/- & 39.28/0.956 &147.9G&0.0595s\\
    DeamNet\cite{ren2021adaptive} &  -/-  & -/- & -/- &-/-&39.47/0.957&582.9G&0.0565s\\
    \midrule
    MPRNet\cite{zamir2021multi} &  \textbf{33.64/0.938} & \textbf{32.91}/\underline{0.916} & \textbf{32.66/0.959} & \textbf{30.96/0.939} &\textbf{39.71/0.958}&565.0G&0.0593s\\
    \midrule
    Res-SL &   33.24/\underline{0.937} & 32.67/\textbf{0.917} & 31.91/\underline{0.957} & 30.17/0.934 & 39.60/\underline{0.957}&\underline{85.0G}&\textbf{0.0356s}\\
    Ufor-SL &   32.96/0.931 & 32.26/0.915 & \underline{32.24}/0.955 & \underline{30.35}/0.933 & \underline{39.61}/0.956&\textbf{57.1G}&0.0540s\\
    DRS-SL &  \underline{33.57}/\textbf{0.938}  & \underline{32.71}/\textbf{0.917} & 31.93/0.956 & 30.26/\underline{0.935} & 39.57/0.956 &132.0G&0.0599s\\
    \bottomrule
  \end{tabular}
\end{table*}

\begin{figure}[t]
\setlength{\abovecaptionskip}{0.1cm}
\setlength{\belowcaptionskip}{-0.6cm}
\centering
  \includegraphics[width=\columnwidth]{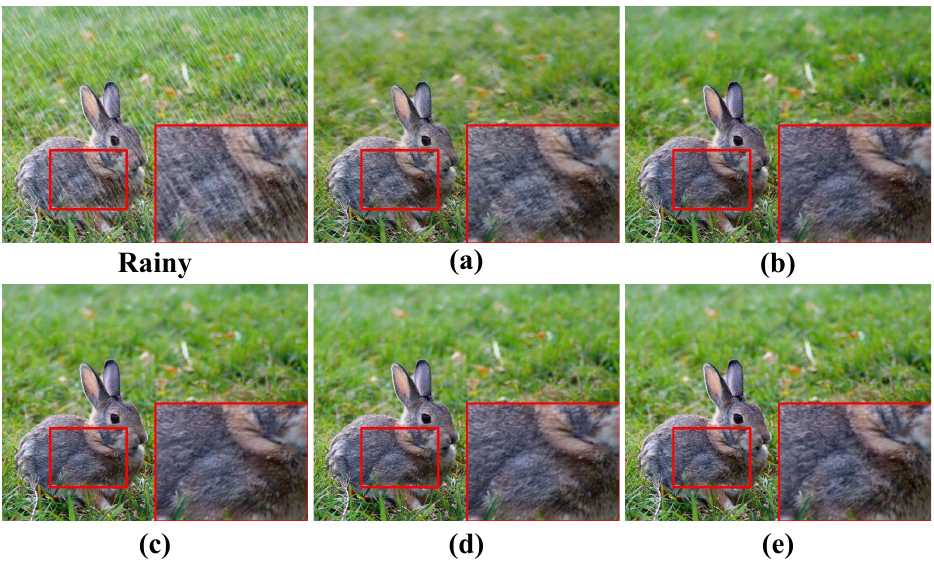}
  \caption{Qualitative ablation study results on Rain1400\cite{fu2017removing}.}
  \label{fig:ablation}
\end{figure}

\subsection{Comparisons with State-of-the-arts}
We conduct a comprehensive comparison of our SLKD strategy against two SOTA image-to-image transfer knowledge distillation methods (DFKD \cite{zhang2021data}, DCKD \cite{fang2023dual}) and seven lightweight SOTA restoration methods (PReNet \cite{ren2019progressive}, RCDNet \cite{wang2020model}, DMPHN \cite{zhang2019deep}, MT-RNN \cite{park2020multi}, VDN\cite{yue2019variational}, DeamNet\cite{ren2021adaptive}, MPR \cite{zamir2021multi}).

\textbf{Knowledge Distillation Methods.} Qualitative comparisons between our strategy and the two SOTA knowledge distillation methods on three restoration tasks are presented in Fig. \ref{fig:first}.
Our strategy excels in removing multiple degradation, producing images that closely match the ground truth
Furthermore, the quantitative results, shown in Table \ref{tab:full-dis}, reveal that our method consistently outperforms others across various teacher models and datasets, demonstrating its effectiveness.

\textbf{Lightweight Methods} We further evaluate our method against SOTA lightweight restoration models. Qualitative and quantitative results are displayed in Fig. \ref{fig:main} and Table \ref{tab:light-dis}, rspectively.
In Fig. \ref{fig:main}, it's evident that our distilled models excel at effectively mitigating multiple degradation, including rain streaks, blur, and noise.
As shown in Table \ref{tab:light-dis}, our distilled models exhibit significantly lower computational complexity compared to other models. 
At the same time, their performance in terms of PSNR \cite{huynh2008scope} and SSIM \cite{wang2004image} is considerably better than that of lightweight models, approaching the performance of the more complex MPRNet \cite{zamir2021multi}.

\begin{table}[tp]
\small
\tabcolsep=7pt
\setlength{\abovecaptionskip}{0.1cm}
\setlength{\belowcaptionskip}{-0cm}
\centering
  \caption{Ablation study results on Rain1400\cite{fu2017removing}.}
  \label{tab:abl}
  \begin{tabular}{ccccc}
    \toprule
    Sets & Source & $L_{NC}$ & $L_{ET}$ & PSNR/SSIM \\
    \midrule
    \multirow{2}{*}{Decoder} & rainy &  &  & 32.20/0.924 \\
     & clean &  &  & 32.61/0.929 \\
    \midrule
    \multirow{2}{*}{Loss}& clean & \checkmark &  & 33.01/0.931 \\
    & clean &  & \checkmark & 32.99/0.932 \\
    \midrule
    Res-SL & clean & \checkmark & \checkmark & \textbf{33.24/0.937} \\
    \bottomrule
  \end{tabular}
\end{table}

\subsection{Ablation Studies}
Ablation studies are presented in Fig. \ref{fig:ablation} and Table \ref{tab:abl}. 
From Fig. \ref{fig:ablation} (a), (b) and corresponding first two rows in Table \ref{tab:abl}, it is evident that using clean images as input for teacher B in IRL, rather than degraded images, significantly enhances the clarity of the restored images.
Additionally, as shown in Fig. \ref{fig:ablation} (b), (c), (d), (e) and corresponding rows 2-5 in Table \ref{tab:abl}, both learning strategies, along with the two losses $L_{NC}$ and $L_{ET}$, independently contribute to the model's ability to remove degradation and reconstruct clean images. 
Together, these learning strategies significantly improve the overall quality of the restored images.

\section{Conclusion}
In this paper, we introduce the \textbf{S}imultaneous \textbf{L}earning \textbf{K}nowledge \textbf{D}istillation (\textbf{SLKD}) framework, designed for compressing image restoration models while preserving performance. 
By integrating \textbf{D}egradation \textbf{R}emoval \textbf{L}earning (\textbf{DRL}) and \textbf{I}mage \textbf{R}econstruction \textbf{L}earning (\textbf{IRL}), SLKD effectively tackles the challenges of compressing image restoration models, allowing the student network to learn from both degraded and clean images.
Experiments across multiple datasets and tasks, deraining, deblurring, and denoising, demonstrate that SLKD significantly reduces model complexity (over 80\% fewer FLOPs and parameters) while maintaining high restoration quality. 
Ablation studies further validate the effectiveness of DRL and IRL in improving image clarity and overall model robustness.

\newpage

\balance

% \bibliographystyle{IEEEtran}
% \bibliography{IEEEabrv, refs}

\end{document}